\documentclass[10pt,twocolumn,letterpaper]{article}

\usepackage{cvpr}              % To produce the CAMERA-READY version
\definecolor{cvprblue}{rgb}{0.21,0.49,0.74}
\usepackage[pagebackref,breaklinks,colorlinks,allcolors=cvprblue]{hyperref}
\usepackage[ruled,vlined,linesnumbered]{algorithm2e}
\usepackage{booktabs}
\usepackage{arydshln}
\usepackage[table]{xcolor}
\usepackage{colortbl}
\usepackage{booktabs}
\usepackage{multirow}
\usepackage{tabularx}

\def\paperID{}
\def\confName{CVPR}
\def\confYear{2026}

\title{Blink: Dynamic Visual Token Resolution for \\ Enhanced Multimodal Understanding}

\author{
    Yuchen Feng$^{1,2}$\thanks{$\quad$ Equal Contribution.}, 
    Zhenyu Zhang$^{3}$\footnotemark[1], 
    Naibin Gu$^{1,2}$, 
    Yilong Chen$^{1,2}$,
    Peng Fu$^{1,2}$\thanks{$\quad$ Corresponding Author.}, 
    Zheng Lin$^{1,2}$, \\
    Shuohuan Wang$^{3}$,
    Yu Sun$^{3}$,
    Hua Wu$^{3}$,
    Weiping Wang$^{1}$,
    Haifeng Wang$^{3}$\\
    \small{$^1$Institute of Information Engineering, Chinese Academy of Sciences,} \\
    \small{$^2$School of Cyber Security, University of Chinese Academy of Sciences, }
    \small{$^3$Baidu Inc.} \\
    {\small \texttt{\{fengyuchen,fupeng\}@iie.ac.cn, zhangzhenyu07@baidu.com}}
}

\begin{document}
\maketitle
\begin{abstract}

% Multimodal large language models (MLLMs) have achieved remarkable progress on vision-language tasks, but still exhibit limited visual perception. Unlike MLLMs, humans dynamically scan and focus on salient regions in a sequential ``blink-like'' process, enabling efficient and adaptive visual understanding. Inspired by this process, we propose \textbf{Blink}, a dynamic perception framework that emulates human-like scanning and focusing within a single forward pass. Specifically, through in-depth experiments,  we identify that MLLMs inherently attend to different visual regions across layers and can enhance visual perception by allocating more computation to salient tokens. Building on this insight, Blink leverages saliency-guided scanning and dynamic token resolution, which estimate token saliency from attention maps and expand or drop tokens via trained amplifiers, called token super-resolution (TokenSR) modules. This dynamic mechanism balances broad exploration and fine-grained focus, thereby enhancing visual perception adaptively and efficiently. Extensive experiments validate Blink, demonstrating its effectiveness in enhancing visual perception and multimodal understanding.
% and its superior accuracy–efficiency balance.

Multimodal large language models (MLLMs) have achieved remarkable progress on various vision-language tasks, yet their visual perception remains limited. Humans, in comparison, perceive complex scenes efficiently by dynamically scanning and focusing on salient regions in a sequential ``blink-like'' process. Motivated by this strategy, we first investigate whether MLLMs exhibit similar behavior. Our pilot analysis reveals that MLLMs naturally attend to different visual regions across layers and that selectively allocating more computation to salient tokens can enhance visual perception. Building on this insight, we propose Blink, a dynamic visual token resolution framework that emulates the human-inspired process within a single forward pass. Specifically, Blink includes two modules: saliency-guided scanning and dynamic token resolution. It first estimates the saliency of visual tokens in each layer based on the attention map, and extends important tokens through a plug-and-play token super-resolution (TokenSR) module. In the next layer, it drops the extended tokens when they lose focus. This dynamic mechanism balances broad exploration and fine-grained focus, thereby enhancing visual perception adaptively and efficiently. Extensive experiments validate Blink, demonstrating its effectiveness in enhancing visual perception and multimodal understanding.

\end{abstract}    
\section{Introduction}
\label{sec:intro}

\begin{figure}[t]
  \centering
  \includegraphics[width=0.82\linewidth]{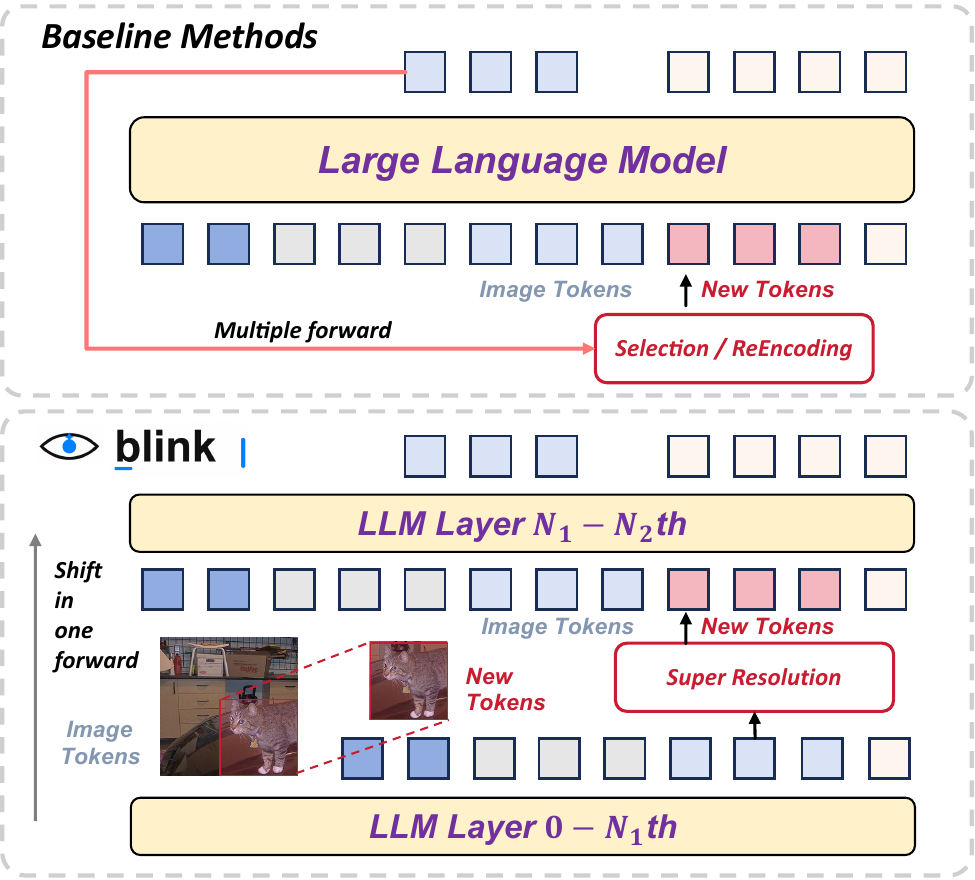}
  \caption{Comparison of the conventional post-hoc paradigm for enhancing salient regions, and our propose Blink. Blink shifts attention across salient regions over layers, achieving dynamic visual perception enhancement.}
  \label{fig:intro}
\end{figure}

Multimodal large language models (MLLMs) have rapidly emerged across various downstream fields, including computer vision and natural language processing~\cite{achiam2023gpt, team2024gemini, bai2025qwen2, chen2024internvl, li2024llava}.  
By coupling a vision encoder with a projector that maps visual features into the language space of the following large language models (LLMs), MLLMs effectively unify visual and textual representations~\cite{alayrac2022flamingo, li2023blip, liu2024improved}. This architecture has demonstrated impressive performance across a wide spectrum of multimodal tasks, including text generation, visual understanding, and cross-modal reasoning.
Despite these advances, MLLMs still struggle with insufficient visual perception capabilities~\cite{wu2312v, cheng2024spatialrgptgroundedspatialreasoning}, often resulting in incorrect answers or hallucinated explanations~\cite{tong2024eyeswideshutexploring, bai2025hallucinationmultimodallargelanguage, jiang2024hallucinationaugmentedcontrastivelearning, favero2024multimodalhallucinationcontrolvisual}. 
Compared with LLMs, MLLMs take additional visual inputs and thus need to reason over both linguistic and visual information, making their performance on multimodal tasks highly dependent on visual perception. 
However, current MLLMs still adopt the conventional LLM architecture for downstream task execution, lack of explicitly enhancing or leveraging salient visual regions~\cite{wu2025groundedchainofthoughtmultimodallarge, chen2025geopqabridgingvisualperception, zhang2025openeyesreasonfinegrained}. Therefore, it is imperative to investigate their visual perception to reveal potential risks in downstream applications.

Unlike naive textual understanding, humans perceive visual scenes through a progressive and dynamic process. They first glance the entire image, then fixate on salient regions, and subsequently shift attention to new areas when the current focus becomes uninformative, as studied in cognitive and visual neuroscience~\cite{kelley2008cortical, maunsell2015neuronal, gmeindl2016tracking, davis2024atlas}. 
This dynamic process, reminiscent of a rapid ``blink'' between salient regions, enables humans to efficiently integrate visual information through sequential attention shifts. 
Inspired by this principle, it is quite promising to equip MLLMs with a similar capability that detects and selects salient regions, adaptively expands attention on them, and shifts focus as needed, thereby progressively guiding the model toward the most informative areas.

% To explore a possible solution, we analyze the visual information processing behavior of MLLMs, yielding two key insights both across and within transformer layers. 
To explore a possible solution, we conduct a pilot analysis of MLLMs’ visual information processing behavior and uncover \textbf{two key insights}, across and within transformer layers.
Across layers, the attention weights vary substantially, suggesting that different layers focus on distinct visual regions with varying attention sharpness. 
Within a layer, we find that increasing computation for high-attention visual tokens, such as replicating salient regions, effectively enhance the visual perception capability. 
\emph{These findings reveal that MLLMs possess an inherent ability to attend to diverse regions across layers and improve visual understanding by allocating more computation to salient tokens}. 
Recent approaches have attempted to identify salient regions by utilizing attention maps or post-hoc selection modules, and improve visual perception through two or more forward pass~\cite{wu2312v, li2025dyfo, zhang2025mllms, yu2025introducing, chen2025inner}.
However, since an image often contain multiple salient regions, these post-hoc methods typically only support focusing on one salient region, thereby limiting both flexibility and efficiency. 
% Therefore, 
It raises a natural question:
\textit{How can we dynamically enhance the visual perception capability of MLLMs within one forward pass?}

Building upon these observations, we propose \textbf{Blink}, a dynamic visual token resolution framework that emulates human-like scanning and focusing within a single forward pass of MLLMs. 
Inspired by the human strategy of alternating between broad visual exploration and fine-grained attention, Blink is designed to dynamically allocate more tokens to salient visual regions, enhancing multimodal understanding without multiple inference steps.
Specifically, it consists of two synergistic modules: saliency-guided scanning and dynamic token resolution. In the inference process, it first scans the attention map to estimate the saliency of visual tokens, and then expands the selected salient regions through a trained amplifier token super-resolution module.
As attention shifts away, the corresponding visual tokens are pruned, allowing a dynamic balance between wide-area scanning and detail focusing.
As shown in Fig.~\ref{fig:intro}, by mimicking this human-like perception strategy, Blink leverages the inherent non-uniformity of attention across transformer layers, enabling adaptive enhancement of visual perception and improving overall model performance.

We conduct experiments on LLaVA-1.5~\cite{liu2024improved} across seven vision-language benchmarks. Our method achieves consistent performance improvement compared with the backbone model, demonstrating notable gains in visual perception and understanding. Even in the variant where the token super-resolution module is replaced with simple training-free interpolation and only the inference pipeline is retained, Blink still surpasses the backbone, confirming the effectiveness of our dynamic framework. Furthermore, comprehensive analyses highlight the contributions of key modules and design choices, while providing insights into the underlying mechanisms of our approach.

In summary, our contributions are threefold:

\begin{itemize}
    \item \textbf{Key Insights.} We reveal two previously overlooked properties of MLLMs, showing that the visual focus naturally shifts across layers and selectively allocating computation to salient regions yields stronger visual perception.
    \item \textbf{Dynamic Framework.} We propose Blink, a dynamic visual token resolution framework that enables adaptive enhancement of multimodal understanding by expanding salient tokens and dropping less informative regions.
    \item \textbf{Empirical Evaluation.} We validate Blink across diverse multimodal benchmarks, demonstrating consistent improvements in visual perception and establishing a simple yet effective plug-in mechanism for enhancing MLLMs.
\end{itemize}

\section{Preliminary}
\label{sec:preliminary}

\subsection{Architecture of MLLMs}

An MLLM typically consists of a vision encoder $F_v(\cdot)$, a projection network $F_p(\cdot)$, and a stack of transformer-based decoder layers that serve as the LLM backbone.
Given an input image $I \in \mathbb{R}^{H \times W \times 3}$, a textual prompt $T$, and a system prompt $S$, the vision encoder first extracts visual features $E_v = F_v(I) \in \mathbb{R}^{h \times w \times d}$, which are projected into the LLM embedding space as $\tilde{E}_v = F_p(F_v) \in \mathbb{R}^{h \times w \times d}$. Meanwhile, the textual prompt with $n$ tokens is embedded as $E_t \in \mathbb{R}^{n \times d}$.
Next, the hidden states are combined as $hs^{(0)} = \operatorname{concat}(\tilde{E}_v, E_t)$, and fed into $L$ decoder layers. Each layer includes a multi-head self-attention (MHA) module and a feed-forward network (FFN) with residual connections. 
% In forward process, the MHA modules capture contextual dependencies by computing attention weights among tokens, while the FFNs apply a non-linear transformation to refine each token representation.
% \begin{equation}
%     \text{Attention}(Q, K, V) = \text{Softmax} \left( \frac{QK^T}{\sqrt{d_k}} + M \right) V,
% \end{equation}
% where $M$ is the causal mask ensuring autoregressive generation. 
% And, the FFN applies a non-linear transformation to each token. Residual connections are applied around both submodules.
% \begin{equation}
%     \text{FFN}(x) = F_\text{down}(\sigma(F_\text{up}(x))),
% \end{equation}
Finally, after all decoder layers, the last hidden states $hs^{(L)}$ are mapped to the vocabulary space via an output projection matrix $W_O \in \mathbb{R}^{d \times V}$, and normalized by Softmax to produce the next-token probabilities. 
In this way, LLM generates tokens in the form of auto-regression based on both visual and textual inputs, integrating multimodal information at each step.

\begin{figure}[t]
  \centering
  \includegraphics[width=\linewidth]{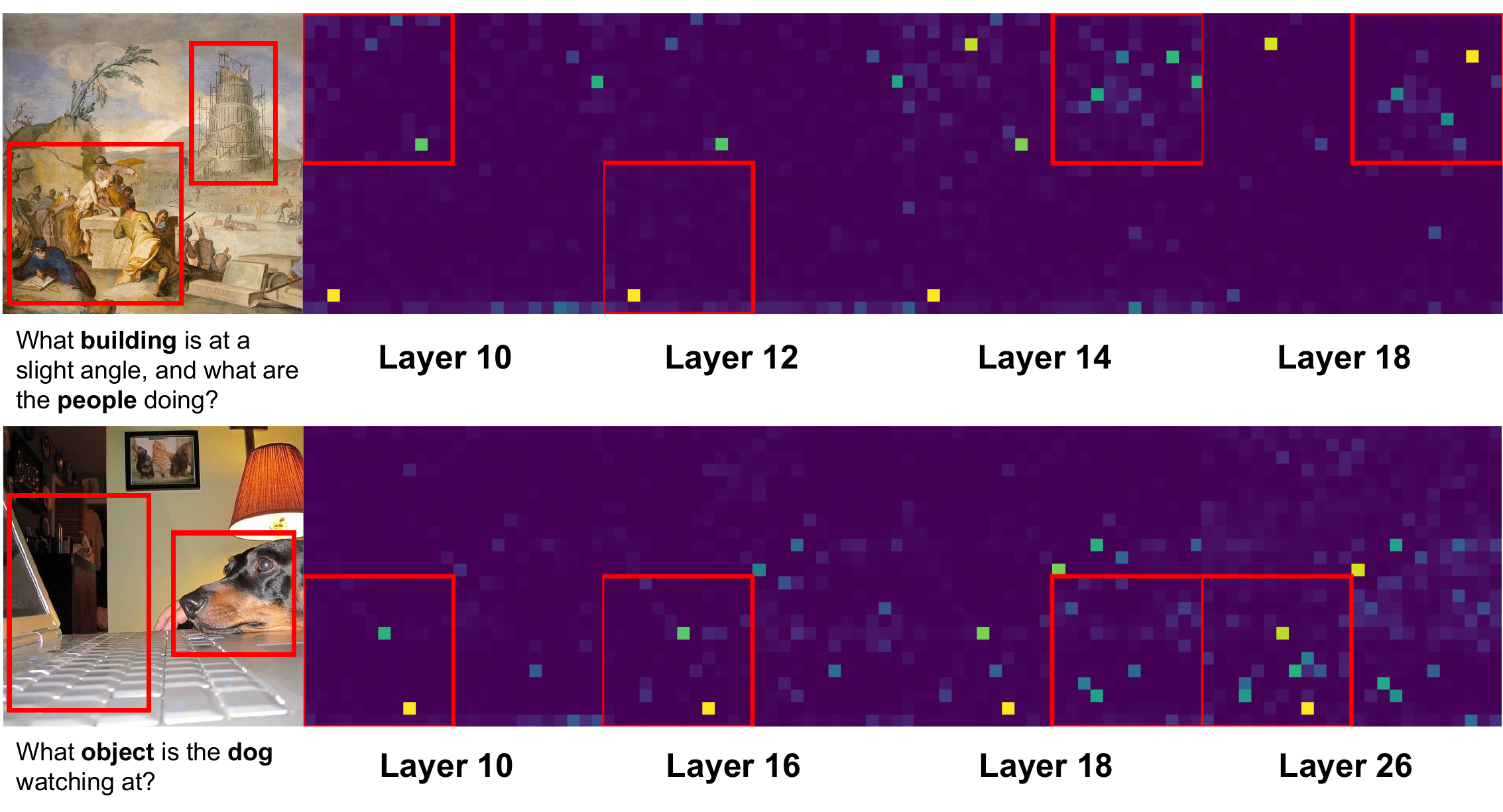}
  \caption{Visualization of attention maps across layers. In the right panels, lighter colors indicate higher attention from the last text token, and red boxes mark the highest-attention regions.}
  \label{fig:insight-1-1}
\end{figure}

\begin{figure}[t]
  \centering
  \includegraphics[width=0.85\linewidth]{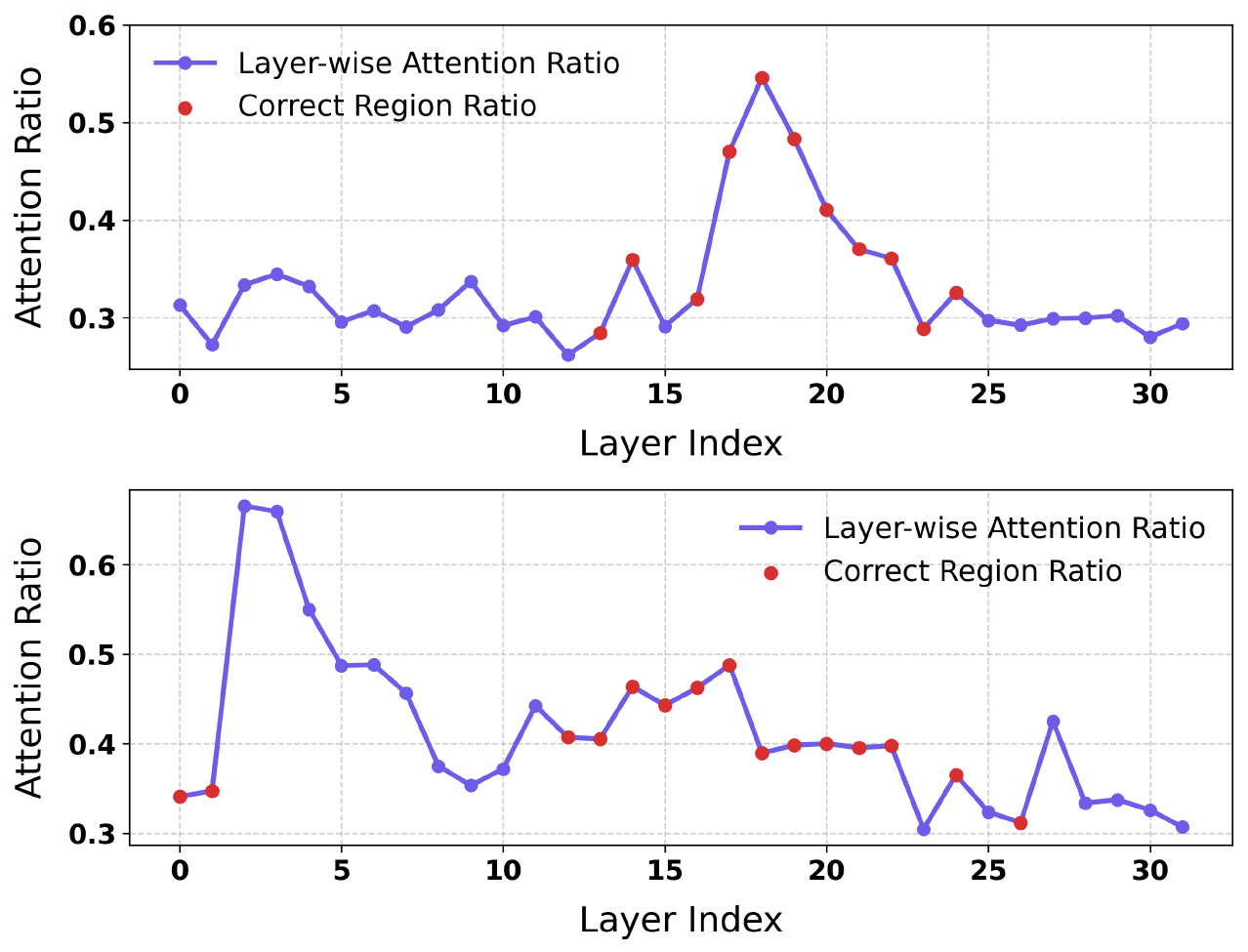}
  \caption{Attention ratio across layers. Purple lines show the ratio of the most attended region to total visual-token attention, and red dots mark layers where the most attended region is correct.}
  \label{fig:insight-1-2}
\end{figure}

\subsection{Key Insights}
\label{sec:insights}

To better understand how MLLMs leverage multimodal information, we take a deep dive into model behaviors both across and within transformer layers. These studies reveal two key insights into how MLLMs perceive and utilize visual information.

\noindent\textbf{Key Insight 1: Attention weights are non-uniformly distributed \underline{across layers}.}
Here, we explore how MLLMs distribute attention over image regions relevant to the text input as the signals propagate across layers. Inspired by prior studies that leverage attention maps to assess token importance~\cite{chen2024image, xing2024pyramiddrop, tu2024vl, ye2025fit, chen2024nacl}, we analyze how attention distributions vary across layers in LLaVA-1.5-7B.

We first examine how attention patterns evolve across layers when processing an image. In Fig.~\ref{fig:insight-1-1}, we treat the final text token as the query and compute attention weights over all visual tokens. For visualization, the attention maps in each layer are reshaped to $H \times W$ to match the spatial layout of the original image, and we highlight the region with the highest cumulative attention weight. 
It is observed that when multiple important targets exist in an image (e.g., the building and people in Case 1, or the object and dog in Case 2), the attention does not remain fixed on a single region. Instead, it progressively shifts its focus over different regions across layers. This behavior indicates that the model inherently possesses the ability to refine its attention progressively and to switch focus among multiple objects.

Furthermore, we analyze how the attention ratio evolves across layers, where the ratio is defined as the attention weight of the most attended region relative to the total visual attention. As shown in Fig.~\ref{fig:insight-1-2}, the purple curve tracks this ratio across transformer depth, and red markers indicate layers where the attended region corresponds to the correct visual area. Two patterns emerge: (1) correct attention tends to occur in the middle layers (approximately layers 12–26), and (2) these layers show notably sharper attention distributions, reflecting greater model confidence when focusing on relevant regions.
Taken together, the above analyses reveal our first insight that different transformer layers exhibit distinct attention pattern, which progressively shift focus across layers and differ in sharpness.

\begin{figure}[t]
  \centering
  \includegraphics[width=0.9\linewidth]{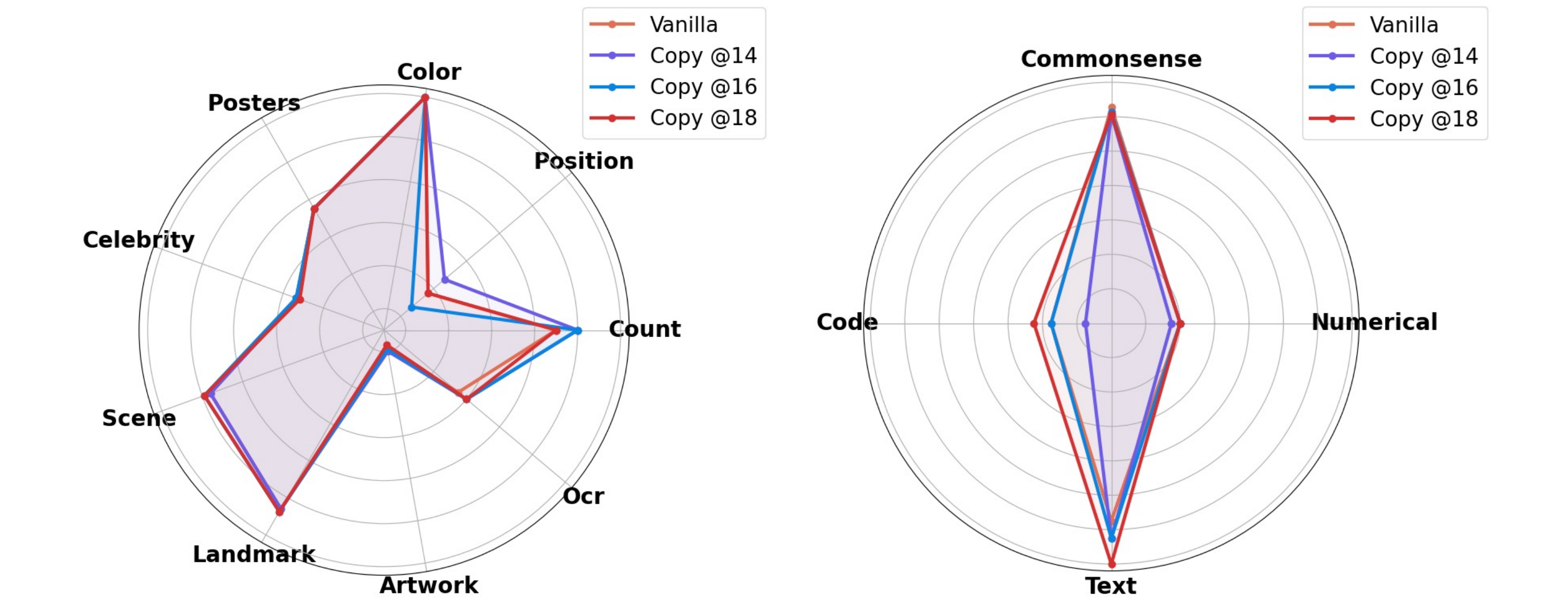}
  \caption{Performance on MME tasks when copying salient visual tokens at different layers. Vanilla denotes the original LLaVA-1.5 performance, and Copy @14/16/18 indicates the layer at which token copying is applied.}
  \label{fig:insight-2}
\end{figure}

\begin{figure*}[t]
  \centering
  \includegraphics[width=0.85\linewidth]{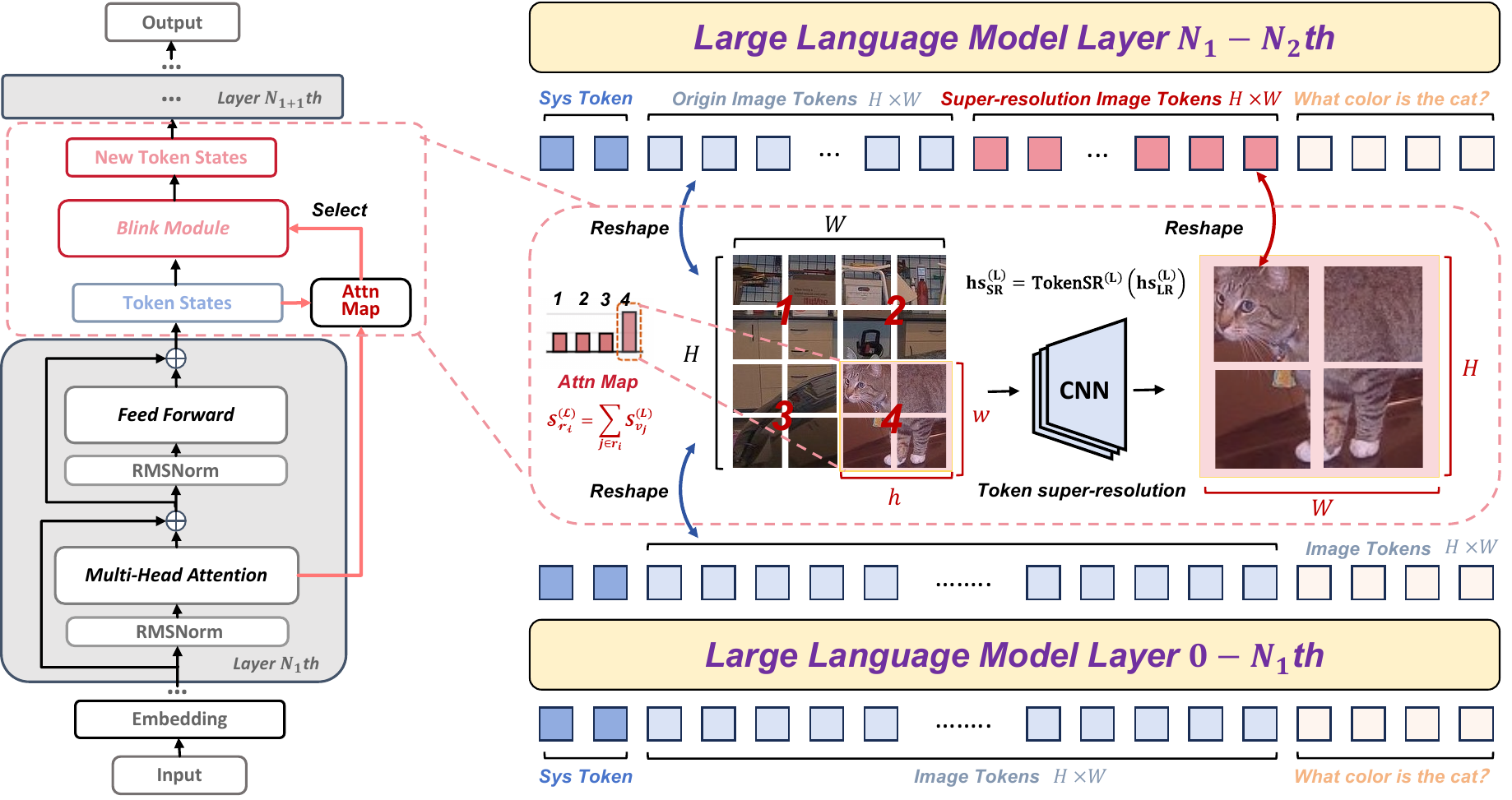}
  \caption{Illustration of our proposed Blink. The left side shows the saliency-guided scanning that determines whether to expand or drop visual tokens across transformer layers. The right side shows within-layer sequence reconstruction, where salient tokens are amplified by the token super-resolution module.}
  \label{fig:overview}
\end{figure*}

\noindent\textbf{Key Insight 2: The computation on salient visual tokens \underline{within a layer} affects visual perception.}
Next, to investigate the effect of computation on salient visual regions within a layer, we copy the visual tokens corresponding to the highest-attention areas at a given layer. These copied tokens are interpolated to match the original sequence length and inserted between the original visual tokens and the text token embeddings in the hidden states. Subsequent layers then process this extended sequence, effectively increasing the computational focus on important regions.

Fig.~\ref{fig:insight-2} shows the evaluation on several perception and cognition tasks in a representative benchmark, MME~\cite{fu2025mmecomprehensiveevaluationbenchmark}. Specifically, it compares the performance of the vanilla LLaVA-1.5 model with that of models where token copying is applied at fixed layers (14, 16, and 18). In the radar plots, a larger enclosed area indicates better task performance, clearly showing that when we copy salient visual tokens at fixed layers, thereby increasing computation on important regions, the performance improves across tasks. These findings form our second insight that allocating more computation to salient visual tokens within a layer can enhance the visual perception capability of MLLMs.

These two insights reveal that MLLMs inherently adjust their visual focus both across and within layers and that dynamically allocating computation to salient regions can further enhance their perception capability. \emph{They motivate us to design a framework exploiting these properties to achieve adaptive visual perception enhancement}.
\section{Method}
\label{sec:method}

In this section, we propose the Blink framework. As shown in Fig.~\ref{fig:overview}, Blink comprises two modules: saliency-guided scanning and dynamic token resolution, simulating human-like rapid scanning and focusing, thereby it can enhance visual perception within a single forward pass in an efficient and adaptive manner.
% In this way, Blink can enhance visual perception within a single forward pass in an efficient and adaptive manner.

\subsection{Saliency-Guided Scanning}

As discussed in Sec.~\ref{sec:insights}, salient visual tokens can be intuitively identified as those receiving higher attention scores. Based on it, we calculate the saliency of each visual token with respect to the last token of the text prompt using. At each involved transformer layer $L$, we denote the hidden states of the visual tokens and the final text token as $hs_v^{(L)}$ and $hs_{t_n}^{(L)}$, respectively. The corresponding key and query representations are obtained through linear projections:
\begin{equation}
\small
    k_v^{(L)} = K_v(hs_v^{(L)}), \quad q_{t_n}^{(L)} = Q_t(hs_{t_n}^{(L)}),
\end{equation}
where $K_v(\cdot)$ and $Q_t(\cdot)$ represent the key and query projection matrices of the MHA module.  
The saliency score $S_v^{(L)}$ of each visual token is then computed as the similarity between $q_{t_n}^{(L)}$ and $k_v^{(L)}$:
\begin{equation}
\small
    S_v^{(L)} = q_{t_n}^{(L)} (k_v^{(L)})^{\!\top}.
\end{equation}

To recover the original spatial structure of the image, we reshape the one-dimensional sequence of visual tokens into a two-dimensional grid of size $H \times W$, corresponding to the spatial layout of the input image. The grid is uniformly divided into $p \times p$ non-overlapping patches of equal size, each with dimensions $h \times w = \left( \frac{H}{p} \right) \times \left( \frac{W}{p} \right)$.  For each patch $r_i$, we compute its aggregated saliency as the sum of token-level scores $\mathcal{S}_{r_i}^{(L)} = \sum_{j \in r_i} S_{v_j}^{(L)}$. The patch with the highest aggregated saliency is identified as the most informative region at layer $L$.

Building on our insight that different layers exhibit distinct attention distributions, including variations in both the most attended regions and the sharpness of these attention peaks, we observe that the model progressively attends to multiple potential regions of interest across layers, with varying degrees of saliency concentration. Following this observation, we further define a saliency ratio between the most salient patch and all visual tokens:
\begin{equation}
\small
    \rho^{(L)} = \frac{\mathcal{S}_{r_{\text{max}}}^{(L)}}{\sum_i \mathcal{S}_{r_i}^{(L)}},
\end{equation}
where $\mathcal{S}_{r_{\text{max}}}^{(L)}$ denotes the saliency of the most informative patch at layer $L$. The saliency ratio $\rho^{(L)}$ quantifies the sharpness of the saliency distribution, reflecting how concentrated the attention is at a specific layer. Higher values of $\rho^{(L)}$ indicate a sharp and localized focus, whereas lower values correspond to a more diffuse and exploratory attention pattern. 
This ratio is further utilized in the following dynamic token resolution module, enabling the model to adaptively refine its visual focus by expanding salient regions for detailed perception and dropping less informative areas, thereby enhancing visual grounding without additional forward passes.
% This saliency ratio is further utilized in dynamic token resolution module to enable adaptive selection of visual objects that require refinement.

\subsection{Dynamic Token Resolution}

% We first define the token resolution as the level of spatial granularity encoded in the hidden states, which reflects how finely each token represents the underlying visual content. 
We define token resolution as the spatial granularity encoded in hidden states, indicating how finely each token represents the underlying visual content.
Based on the saliency ratio $\rho^{(L)}$, we adopt an adaptive thresholding strategy to dynamically determine whether the model should expand or drop visual tokens at each layer, thereby achieving adjustment of token resolution. Specifically, when $\rho^{(L)} > \tau_{\text{exp}}$, the model activates the token super-resolution module to enhance and expand tokens within salient regions, where $\tau_{\text{exp}}$ is a predefined threshold. On the contrary, when $\rho^{(L)} < \tau_{\text{drop}}$, redundant tokens are discarded to reduce computational overhead.
Next, we provide a detailed description of how the expansion and dropping operations are respectively applied to the corresponding visual tokens.

\subsubsection{Token Expansion}

\noindent\textbf{Token super-resolution module.}
Inspired by image super-resolution techniques~\cite{dong2015image, dong2016accelerating, kim2016accurate}, we introduce a trainable token amplifier, termed the token super-resolution (TokenSR) module, to refine visual token representations. At each decoder layer $L$, for low-resolution visual tokens $hs_{LR}^{(L)}$ selected for expansion, the module reconstructs a higher-resolution token sequence $hs_{SR}^{(L)}$ that captures finer local semantics and structural details:
\begin{equation}
\small
hs_{SR}^{(L)} = \text{TokenSR}^{(L)}(hs_{LR}^{(L)}).
\end{equation}
Each TokenSR module consists of three convolutional layers with ReLU activations, where each $\text{Conv}_{k}$ denotes a 2D convolutional layer with $c_k$ filters of kernel size $f_k \times f_k$. This hierarchical structure allows the module to refine local details, while keeping the overall semantic information consistent across layers.
During training, each TokenSR module receives the hidden states of salient regions from a full image $I_{\text{full}}$ and produces low-resolution tokens $hs_{SR}^{(L)}$. The corresponding cropped image $I_{\text{crop}}$ provides reference hidden states $hs_{\text{crop}}^{(L)}$. The module is trained by minimizing the KL divergence between enhanced and reference tokens. 
This supervision encourages the TokenSR module to reconstruct high-resolution token representations for salient regions that match the fine-grained structures of the cropped images. Only the parameters of the TokenSR module are updated, while the MLLM backbone remains frozen.
% \begin{equation}
%     \mathcal{L}_{\text{TokenSR}} = D_{\mathrm{KL}}\big(hs_{\text{crop}}^{(L)} \,\|\, hs_{SR}^{(L)}\big).
% \end{equation}

% The TokenSR module is trained to produce high-resolution tokens aligned with cropped images, with the backbone frozen.

\noindent\textbf{Sequence reconstruction.}
Within the token expansion process, the enhanced tokens produced by the TokenSR module at decoder layer $L$ are directly inserted between the original visual tokens and the text tokens. For each sample $i$, salient patch tokens $hs_{\text{patch}}^{(L,i)} \in \mathbb{R}^{h \times w \times d}$ are first extracted from the full-image hidden states, reshaped into a 2D feature map, interpolated to $H \times W$ via bilinear interpolation and processed by the TokenSR module to obtain enhanced tokens $hs_{\text{SR}}^{(L,i)} \in \mathbb{R}^{H \times W \times d}$. These enhanced tokens are then flattened into a 1D sequence. The final output sequence of the corresponding decoder layer is constructed as:
\begin{equation}
\small
    hs_i^{(L,\text{output})} = [\, hs_{s}^{(L,i)}; \, hs_{v}^{(L,i)};\, hs_{\text{SR}}^{(L,i)};\, hs_{t}^{(L,i)} \,].
\end{equation}
The attention mask and positional embeddings are updated to accommodate the expanded token sequence, ensuring that the transformer correctly attends to the enhanced visual regions while maintaining causal consistency.

\subsubsection{Token Drop}

Complementary to token expansion, the token drop mechanism prunes low-saliency visual tokens. If saliency ratio $\rho^{(L)}$ falls below $\tau_{\text{drop}}$, the corresponding $hs_{\text{SR}}^{(L)}$ is removed. And, the sequence is reverted to its original form of:
\begin{equation}
\small
    hs_i^{(L,\text{output})} = [\, hs_{s}^{(L,i)}; \, hs_{v}^{(L,i)};\, hs_{t}^{(L,i)} \,].
\end{equation}
This operation effectively suppresses over-attended yet uninformative areas, preventing weakly supported salient regions from biasing inference and helping subsequent layers focus on more reliable visual evidence.
% allowing subsequent layers to concentrate on more salient regions.

% All the above computations are performed before the layer normalization of the corresponding transformer layer.
% All dynamic token operations, including both expansion and drop, are performed before the layer normalization of the corresponding transformer layer.

\begin{table*}[t]
    \centering
    \small
    \setlength{\tabcolsep}{4pt}
    \renewcommand{\arraystretch}{1.2}
    \begin{tabular}{p{2.5cm} *{11}{c}}
        \toprule
        \multirow{2}{*}{\textbf{Method}} &
        \multicolumn{11}{c}{\textbf{MME$_{\text{Perception}}$}} \\
        \cmidrule(lr){2-12}
        & Exist. & Count & Pos. & Color & Poster & Celeb. & Scene & Landm. & Artw. & OCR & Total \\
        \midrule
        Vanilla & 190.00 & 155.00 & 128.33 & 170.00 & \textbf{147.62} & \textbf{136.76} & \textbf{158.00} & \textbf{163.00} & 119.50 & \textbf{137.50} & 1505.72 \\
        \rowcolor{blue!5}
        \textit{Blink-interp (Ours)} & \textbf{190.00} & \textbf{160.00} & 133.33 & \textbf{170.00} & \textbf{147.62} & 135.88 & 157.25 & \textbf{163.00} & 119.50 & \textbf{137.50} & 1514.08 \\
        \rowcolor{blue!5}
        \textit{Blink (Ours)} & \textbf{190.00} & \textbf{160.00} & \textbf{138.33} & \textbf{170.00} & 147.28 & 135.88 & \textbf{158.00} & \textbf{163.00} & \textbf{119.75} & \textbf{137.50} & \textbf{1519.74} \\
        
        \midrule
        
        \multirow{2}{*}{\textbf{Method}} &
        \multicolumn{5}{c}{\textbf{MME$_{\text{Cognition}}$}} & \multirow{2}{*}{\textbf{GQA}} & \multirow{2}{*}{\textbf{MMBench}} & \multirow{2}{*}{\textbf{MMBench$_{\text{CN}}$}} & \multirow{2}{*}{\textbf{POPE}} & \multirow{2}{*}{\textbf{SQA$_{\text{Img}}$}} & \multirow{2}{*}{\textbf{MM-Vet}} \\
        \cmidrule(lr){2-6}
        & CS & Num & Text & Code & Total & & & & & & \\
        \midrule
        Vanilla & 112.86 & 70.00 & \textbf{107.50} & \textbf{67.50} & 357.86 & 61.93 & 64.60 & 58.08 & 85.17 & 69.46 & 32.20 \\
        \rowcolor{blue!5}
        \textit{Blink-interp (Ours)} & 110.71 & 70.00 & \textbf{107.50} & 65.00 & 353.21 & 61.93 & \textbf{64.69} & 58.51 & 85.17 & 69.51 & 31.70 \\
        \rowcolor{blue!5}
        \textit{Blink (Ours)} & \textbf{114.29} & \textbf{75.00} & \textbf{107.50} & 65.00 & \textbf{361.79} & \textbf{61.98} & \textbf{64.69} & \textbf{58.59} & \textbf{85.23} & \textbf{69.66} & \textbf{33.40} \\
        \bottomrule
    \end{tabular}

    \caption{Downstream task performance across multiple benchmarks on LLaVA-1.5. Results of other backbones are provided in Appendix~\ref{sec:appendix_backbone}. Vanilla denotes the base model, and our methods correspond to two configurations of Blink, where -interp indicates the variant that replaces the amplifier with training-free interpolation while retaining the Blink inference pipeline. The best scores are in \textbf{bold}.}
    \label{tab:downstream}
\end{table*}

\subsection{Workflow of Blink}

% Overall, the complete procedure is presented in Alg.~\ref{alg:blink}. 
% The workflow of Blink integrates the two main stages into a unified procedure that adaptively adjusts the visual token resolution within selected decoder layers. 

% All operations are performed before the layer normalization of each corresponding transformer layer, ensuring that the transformer can process dynamically expanded or pruned tokens without modifying the backbone computations.

Overall, the procedure of Blink is summarized in Alg.~\ref{alg:blink}. 
Blink integrates its two main stages into a unified workflow that adaptively adjusts the visual token resolution within selected decoder layers. By first identifying the most informative visual patches and then selectively expanding or dropping tokens based on saliency, the framework mimics human-like visual attention patterns.
All operations are performed before the layer normalization of each corresponding transformer layer, ensuring that the transformer can process dynamically expanded or pruned tokens without modifying the backbone computations.

\begin{algorithm}[t]
\caption{Workflow of Blink}\label{alg:blink}
\small
\SetAlgoNlRelativeSize{0}
\SetInd{0.8em}{0.8em} % 
\KwIn{MLLM $\mathcal{M}$, selected layers $\mathcal{L}_{sel}$, thresholds $\tau_{\text{exp}}, \tau_{\text{drop}}$, number of divided patches $p \times p$}
\KwOut{Updated hidden states $\{hs^{(L,\text{output})}\}$}

\For{$L \in \mathcal{L}_{sel}$}{
    $\triangleright$ \textbf{Saliency-Guided Scanning}\;
    Reshape $hs_v^{(L)}$ into $H \times W$ grid and divide into $p$ patches\;
    Compute patch saliency $\mathcal{S}_{r_i}^{(L)}$ and saliency ratio $\rho^{(L)}$\;
    
    $\triangleright$ \textbf{Dynamic Token Resolution}\;
    \eIf{$\rho^{(L)} > \tau_{\text{exp}}$}{
        $\triangleright$ \textsc{Expand}: $hs_{SR}^{(L)} \gets \textsc{TokenSR}(hs_{\text{patch}}^{(L)})$\;
        $hs^{(L,\text{output})} \gets [hs_s^{(L)}; hs_v^{(L)}; hs_{SR}^{(L)}; hs_t^{(L)}]$\;
    }{
        \eIf{$\rho^{(L)} < \tau_{\text{drop}}$}{
            $\triangleright$ \textsc{Drop}: remove $hs_{SR}^{(L)}$, revert to original sequence\;
            $hs^{(L,\text{output})} \gets [hs_s^{(L)}; hs_v^{(L)}; hs_t^{(L)}]$\;
        }{
            $\triangleright$ keep $hs^{(L)}$ unchanged\;
        }
    }
    
    $\triangleright$ update Attn\_mask and Position\_ids\;
}
\Return{$\{hs^{(L,\text{output})}\}$}
\end{algorithm}

\begin{table*}[t]
    \centering
    \small
    \setlength{\tabcolsep}{6pt}
    \renewcommand{\arraystretch}{1.15}
    \begin{tabular}{l l l cc ccc}
        \toprule
        \multicolumn{2}{l}{\multirow{2}{*}{\textbf{Dataset}}} &
        \multirow{2}{*}{\textbf{Method}} &
        \multicolumn{2}{c}{\textbf{Key Modules}} &
        \multicolumn{3}{c}{\textbf{Ratio Thresholds}} \\
        \cmidrule(lr){4-5} \cmidrule(lr){6-8}
        \multicolumn{2}{l}{} & &
        w/o SGS & w/o DTR & w/o Drop & 
        High $\tau_{\text{exp}}$ / Low $\tau_{\text{drop}}$ & High $\tau_{\text{exp}}$ \\

        \midrule
        & & \cellcolor{blue!5}\textit{Blink-interp (Ours)} & & & & & \\
        \multirow{3}{*}{\textbf{MME}} &
        \multicolumn{1}{|l}{Perp.} & \cellcolor{blue!5}1514.08 & 1509.83 & 1514.73 & 1514.83 & 1509.83 & 1509.83 \\
        & \multicolumn{1}{|l}{Cong.} & \cellcolor{blue!5}353.21 & 355.36 & 343.93 & 351.07 & 355.71 & 357.86\\
        & \multicolumn{1}{|l}{Total} & \cellcolor{blue!5}1867.29
        & 1865.19\textsubscript{\textcolor[HTML]{d63031}{- 2.10}}
        & 1858.66\textsubscript{\textcolor[HTML]{d63031}{- 8.63}}
        & 1865.90\textsubscript{\textcolor[HTML]{d63031}{- 1.39}}
        & 1865.54\textsubscript{\textcolor[HTML]{d63031}{- 1.75}}
        & 1867.69\textsubscript{\textcolor[HTML]{0984e3}{+ 0.40}} \\

        \midrule
        & & \cellcolor{blue!5}\textit{Blink (Ours)} & & & & & \\
        \multirow{3}{*}{\textbf{MME}} & 
        \multicolumn{1}{|l}{Perp.} & \cellcolor{blue!5}1519.74 & 1519.74 & 1478.67 & 1524.74 & 1509.83 & 1509.83 \\
        & \multicolumn{1}{|l}{Cong.} & \cellcolor{blue!5}361.79 & 359.64 & 361.79 & 359.29 & 357.86 & 355.71 \\
        & \multicolumn{1}{|l}{Total} & \cellcolor{blue!5}1881.53
        & 1879.38\textsubscript{\textcolor[HTML]{d63031}{- 2.15}} 
        & 1840.46\textsubscript{\textcolor[HTML]{d63031}{- 41.07}} 
        & 1884.03\textsubscript{\textcolor[HTML]{0984e3}{+ 2.50}} 
        & 1867.69\textsubscript{\textcolor[HTML]{d63031}{- 13.84}} 
        & 1865.54\textsubscript{\textcolor[HTML]{d63031}{- 15.99}} \\
        
        \bottomrule
    \end{tabular}
    \caption{Performance of Blink and its interpolation variant on the MME benchmark across key modules (SGS, DTR) and ratio threshold settings. Rows labeled w/o SGS and w/o DTR show results with the corresponding module removed. High $\tau_{\text{exp}}$ / Low $\tau_{\text{drop}}$ denotes high expansion threshold with low drop threshold, while High $\tau_{\text{exp}}$ denotes high expansion threshold with normal drop threshold.}
    \label{tab:ablation}
\end{table*}

\section{Experiments}
\label{sec:exp}

\subsection{Experimental Settings}

% \noindent\textbf{Evaluation and Dataset.} 

We evaluate our proposed Blink on LLaVA-1.5-7B~\cite{liu2024improved} across seven commonly used multimodal datasets, including MME~\cite{fu2025mmecomprehensiveevaluationbenchmark}, GQA~\cite{hudson2019gqa}, MMBench, MMBench-CN~\cite{liu2024mmbench}, POPE~\cite{li2023evaluating}, ScienceQA~\cite{lu2022learn}, and MM-Vet~\cite{yu2023mm}. We evaluate Blink and its variant Blink-interp, where the trainable modules are replaced with bilinear interpolation, while retaining the Blink inference pipeline. For training the token super-resolution modules, we use the processed LLaVA-1.5 training data, which is a composite dataset consisting of COCO~\cite{lin2014microsoft}, GQA~\cite{hudson2019gqa}, OCR-VQA~\cite{mishra2019ocr}, TextVQA~\cite{singh2019towards}, and VisualGenome~\cite{krishna2017visual}. For each image, we crop it into four quadrants and record their positions relative to the original image. 
% During training, the model takes pairs of full images and cropped quadrants, enabling supervision that aligns the expanded tokens with their corresponding positions in the original image. 
Detailed training hyper-parameters and inference settings can be found in Appendix~\ref{sec:appendix_train} and ~\ref{sec:appendix_infer}.

\subsection{Main Results}

Tab.~\ref{tab:downstream} presents the downstream performance across seven tasks. Our proposed Blink consistently improves total scores of MME$_{\text{Perception}}$ and MME${_\text{Cognition}}$ over the base model, with gains of 14.02 and 3.93, respectively. Additionally, Blink achieves the highest scores on GQA, MMBench, MMBench-CN, POPE, ScienceQA, and MM-Vet. These results demonstrate the impact of Blink in enhancing multimodal understanding and visual perception.

Blink-interp increases the total score of MME${_\text{Perception}}$ by 8.36, and despite underperforming on MME${_\text{Cognition}}$ and MM-Vet, it surpasses or matches the base model on all other benchmarks. Moreover, the fully trained Blink outperforms Blink-interp on all benchmarks except for MMBench, where Blink-interp achieves comparable performance. These results highlight both the effectiveness of the dynamic inference pipeline and the robustness of the token super-resolution module.

\begin{figure}[t]
  \centering
  \includegraphics[width=\linewidth]{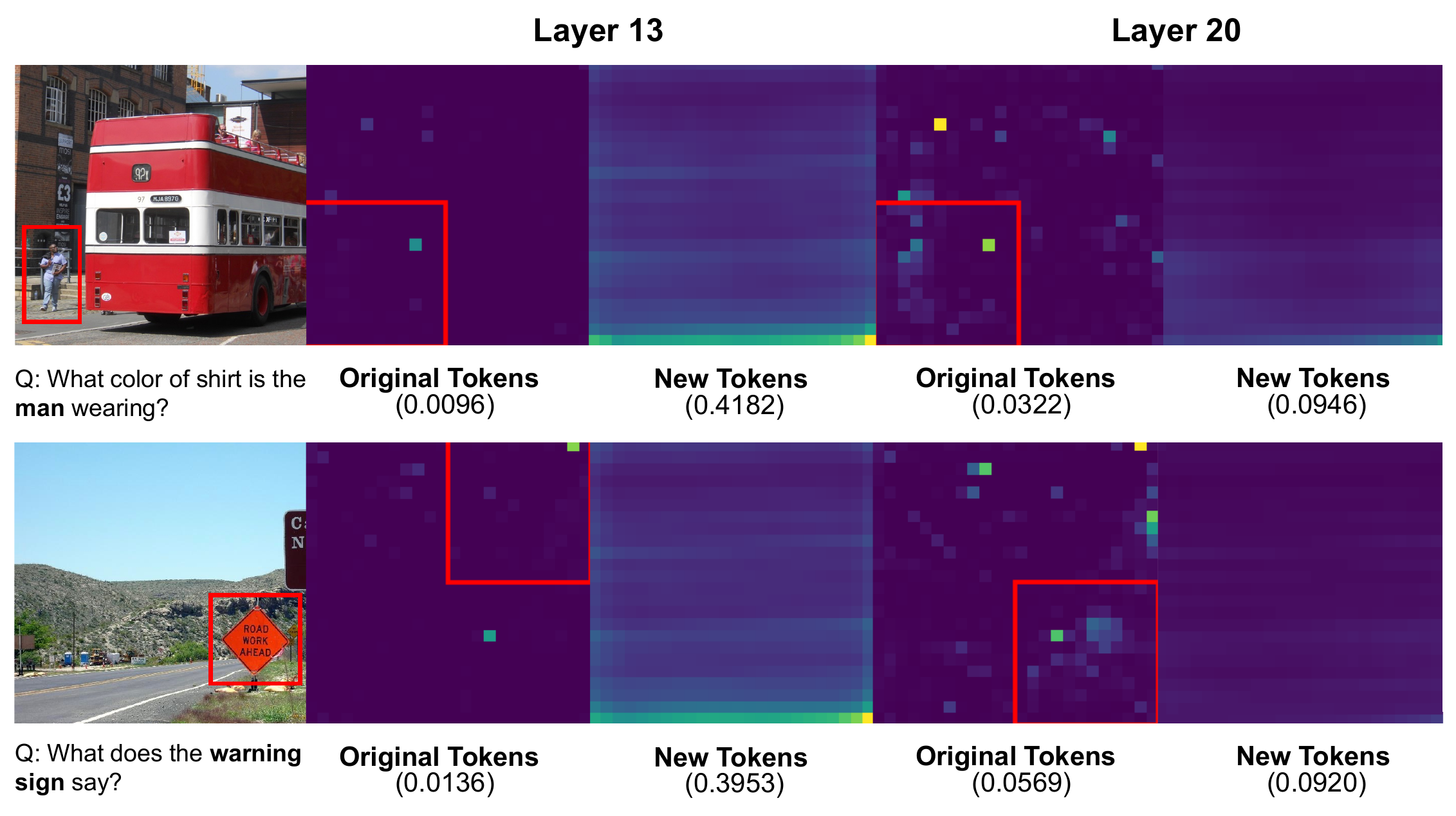}
  \caption{Visualization of attention redistribution after token expansion. Red boxes in the original image indicate the ground-truth important regions. The right panels show attention distributions on the original and expanded visual tokens at layers 13 and 20. See Appendix~\ref{sec:appendix_case} for more visualization results.}
  \label{fig:further-1}
\end{figure}

\subsection{Ablation Studies}

% \subsubsection{Key Modules}
\noindent\textbf{Key Modules.} 
To validate the effectiveness of our proposed modules, we perform ablation studies using MME benchmark on the two key components of Blink: saliency-guided scanning (SGS) and dynamic token resolution (DTR). In the w/o SGS variant, token selection for expansion or drop is performed randomly, without guidance of saliency maps. In the w/o DTR variant, token operations follow a fixed expand–drop cycle across layers, without dynamically adjusting based on thresholds. As shown in Tab.~\ref{tab:ablation}, removing either SGS or DTR results in notable performance drops for both Blink and its variant Blink-interp. Specifically, removing SGS results in the total score dropping by 2.10 for Blink-interp and 2.15 for Blink, and removing DTR causes a larger decrease of 8.63 for Blink-interp and 41.07 for Blink, underscoring the critical role of both saliency-guided scanning and dynamic token resolution in effective token selection and capturing salient visual information.

% \subsubsection{Ratio Thresholds}
\noindent\textbf{Ratio Thresholds.} 
We further investigate the impact of different expansion and drop thresholds on performance. Three alternative settings to our main configuration are considered: (1) the extreme case of disabling token drop (w/o Drop), (2) high expansion threshold and low drop threshold (High $\tau_{\text{exp}}$ / Low $\tau_{\text{drop}}$), and (3) high expansion threshold only (High $\tau_{\text{exp}}$), with the drop threshold kept at its default value. As shown in Tab.~\ref{tab:ablation}, these settings lead to varying performance compared to our default configuration. Specifically, for Blink-interp, the High $\tau_{\text{exp}}$ setting yields only a marginal improvement of 0.40 in the total score, whereas the w/o Drop and High $\tau_{\text{exp}}$ / Low $\tau_{\text{drop}}$ settings reduce performance by 1.39 and 1.75, respectively. For Blink, disabling token drops increases performance by 2.50, while the High $\tau_{\text{exp}}$ / Low $\tau_{\text{drop}}$ and High $\tau_{\text{exp}}$ settings decrease it by 13.84 and 15.99. These results indicate that completely disabling token deletion leads to divergent outcomes, while excessively high expansion thresholds limit effective token expansion, demonstrating that both token expansion and token drop contribute to our framework.

\begin{table}[t]
    \centering
    \small
    \setlength{\tabcolsep}{6pt}
    \renewcommand{\arraystretch}{1.2}
    \begin{tabular}{l c c c}
        \toprule
        \textbf{Layers} & \textbf{Perc.} & \textbf{Cogn.} & \textbf{Total} \\
        \midrule
        11--17 & 
        1518.20\textsubscript{\textcolor[HTML]{d63031}{- 1.54}} &
        364.29\textsubscript{\textcolor[HTML]{0984e3}{+ 2.50}} &
        1882.49\textsubscript{\textcolor[HTML]{0984e3}{+ 0.96}} \\
        
        \rowcolor{blue!5}
        \textit{12--18 (Ours)} & 1519.74 & 361.79 & 1881.53 \\
        
        13--19 & 
        1511.97\textsubscript{\textcolor[HTML]{d63031}{- 7.77}} &
        362.14\textsubscript{\textcolor[HTML]{0984e3}{+ 0.35}} &
        1874.11\textsubscript{\textcolor[HTML]{d63031}{- 7.42}} \\
        
        14--20 & 
        1512.06\textsubscript{\textcolor[HTML]{d63031}{- 7.68}} &
        362.14\textsubscript{\textcolor[HTML]{0984e3}{+ 0.35}} &
        1874.20\textsubscript{\textcolor[HTML]{d63031}{- 7.33}} \\
        
        \bottomrule
    \end{tabular}
    \caption{Performance of Blink on the MME benchmark with different ranges of transformer layers. 12--18 denotes the layer range used in our main configuration, and other ranges correspond to alternative settings. The best scores are in \textbf{bold}.}
    \label{tab:further-2-1}
\end{table}

\subsection{Analysis of Attention Redistribution}
\label{sec:exp_attn}

In this section, we investigate how tokens generated by the token super-resolution module influence attention redistribution across transformer layers after expansion. This helps verify whether the expanded tokens guide the model to focus better and improve visual perception. For clarity, expansion is applied at layer 12, and the new tokens are retained in all subsequent layers.

Fig.~\ref{fig:further-1} shows attention maps at different layers to illustrate this spatial redistribution. In both cases, attention over the new tokens is more evenly distributed and consistently higher than that of the original tokens, indicating effective focus. In Case 2, attention initially concentrates on an incorrect region at layer 13 but shifts to the correct region by layer 20, consistent with Sec.~\ref{sec:insights}, where attention is progressively refined across layers.

\subsection{Analysis of Layer Range}

We observe that although selecting intermediate layers (e.g., 12--18) generally yields strong performance, small shifts in the layer range can lead to noticeable differences. As shown in Tab.~\ref{tab:further-2-1}, shifting from 12--18 to 11--17 slightly improves the total score by 0.96, while shifting to 13--19 or 14--20 decreases performance by 7.42 and 7.33, respectively. This suggests that different layer ranges contribute unevenly, and even minor changes can have non-negligible impact.

\begin{figure}[t]
  \centering
  \includegraphics[width=0.85\linewidth]{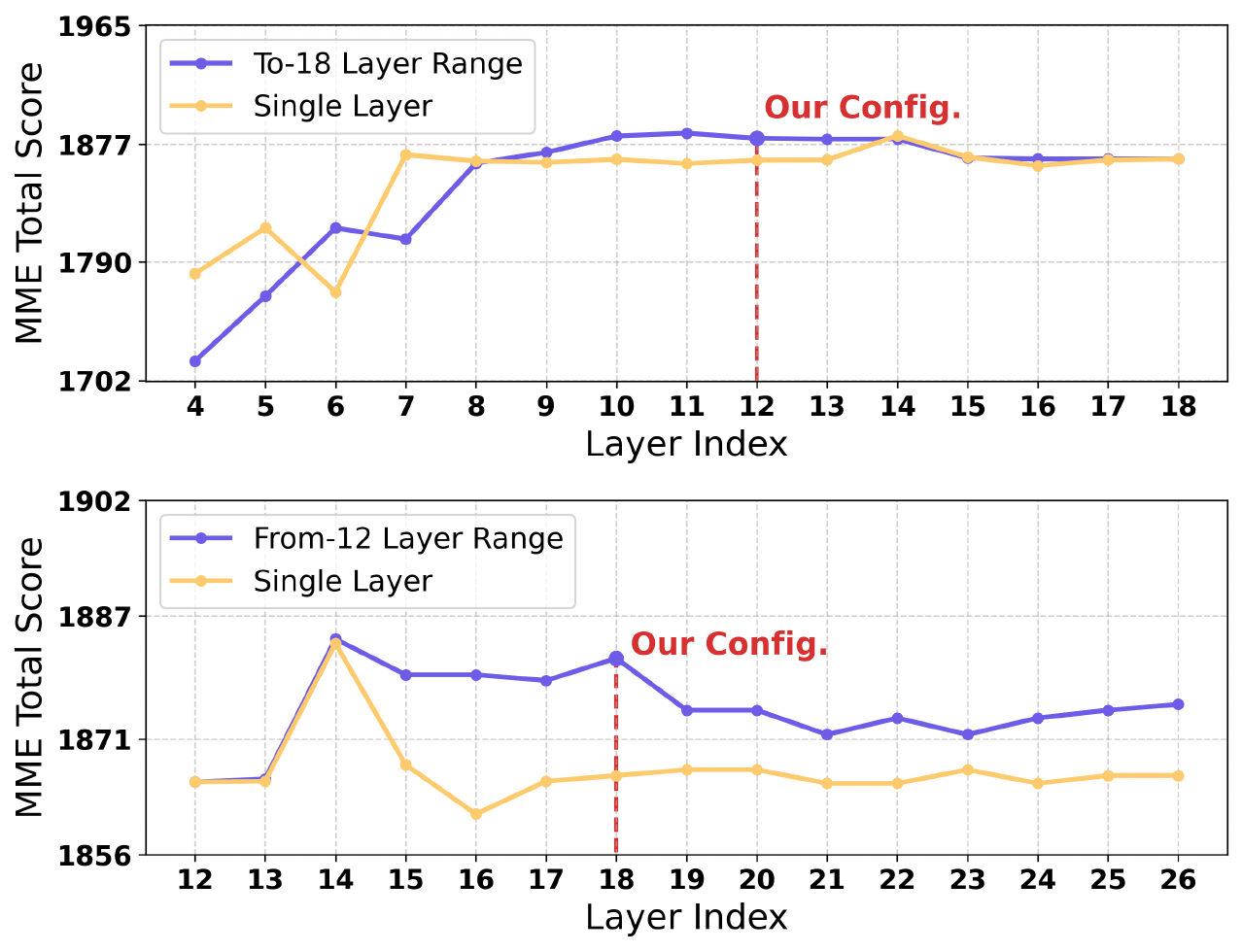}
  \caption{Performance changes on the total MME score with different layer ranges. The top panel varies the start layer (up to 18), and the bottom panel varies the end layer (from 12). Purple lines show performance across ranges, and yellow lines show single-layer dynamic performance.}
  \label{fig:further-2}
\end{figure}

To better understand this effect, we further evaluate the importance and contribution of each layer through three analyses: (1) varying the starting layer with the ending layer fixed at 18, (2) varying the ending layer with the starting layer fixed at 12, and (3) applying dynamic token operations to individual layers. As shown in Fig.~\ref{fig:further-2}, single-layer selection leads to clear performance differences, indicating that attention and accuracy vary across layers. In particular, layers around 7--26 produce much stronger improvements, whereas using only early layers around 4--6 leads to substantial performance drops. Moreover, the choice of layer range affects performance. In the first analysis, excluding layers 4--9 yields better results, whereas in the second analysis, including layers 14--18 causes a noticeable drop. These observations are consistent with Sec.~\ref{sec:insights}, where correct attention tends to emerge in the middle layers. And, from the curves marked with our config, it can be seen that we adopt a broadly effective layer range, though more fine-grained layer selection may further improve performance. 
% \footnote{More analysis experiments, future directions, and limitations are provided in the supplementary materials.}

\section{Related Work}
\label{sec:related}

\noindent\textbf{Multimodal Large Language Models.}
Driven by the success of large language models (LLMs), increasing attention has been devoted to developing end-to-end multimodal large language models (MLLMs). Early works such as CLIP~\cite{radford2021learning} and ALIGN~\cite{jia2021scaling} aligned visual and textual representations through contrastive learning, laying the foundation for cross-modal understanding. Following this paradigm, MLLMs typically employ a vision encoder and a projector to map visual embeddings into the language space of an LLM. Representative models such as BLIP-2~\cite{li2023blip}, LLaVA~\cite{liu2023visual}, InternVL~\cite{chen2024internvl} and the Qwen-VL~\cite{bai2023qwenvlversatilevisionlanguagemodel} series adopt this modular design, where projected visual tokens and textual embeddings are jointly processed by the LLM for multimodal tasks. However, this unified token-based paradigm overlooks modality differences, leading to limited visual perception and hallucination issues~\cite{tong2024eyeswideshutexploring, bai2025hallucinationmultimodallargelanguage, jiang2024hallucinationaugmentedcontrastivelearning, favero2024multimodalhallucinationcontrolvisual}. Consequently, performance on multimodal reasoning heavily depends on accurate visual grounding~\cite{wu2312v, cheng2024spatialrgptgroundedspatialreasoning}, motivating further improvements in visual perception.

\noindent\textbf{Visual Perception Enhancement of MLLMs.}
To address these limitations, recent works have explored enhancing the visual perception of MLLMs through region selection, feature refinement, or adaptive attention, dynamically adjusting visual focus without retraining the backbone. V*~\cite{wu2312v} introduces a guided visual search mechanism that leverages LLM knowledge for efficient querying and collaborative reasoning. 
% By integrating V* with an MLLM, the model can perform contextual understanding and precise targeting of specific visual elements, highlighting the importance of incorporating visual search capabilities in multimodal systems.
DyFo~\cite{li2025dyfo} employs a dynamic focusing module to refine spatial attention on task-relevant regions. Other methods improve fine-grained perception by cropping, recomposing, or re-encoding image regions. For example, ViCrop~\cite{zhang2025mllms} identifies informative patches via attention maps from fixed layers, while Visual Perception Token~\cite{yu2025introducing} enhances focus through region-level cropping and re-encoding, and other works~\cite{gu2025beamlora, feng2025dive} perform parameter-level expansion. Although effective, these approaches generally rely on fixed-layer attention or post-hoc modules and often require multiple forward passes, preventing adaptive refinement of regions of interest within a single inference and limiting both flexibility and efficiency.
\section{Conclusion}
\label{sec:conclusion}

In this work, we propose Blink, a dynamic visual token resolution framework that emulates human-like scanning and focusing within a single forward pass of MLLMs. Inspired by our two key insights on the visual attention shifts across layers and the benefits of allocating more computation to salient regions, Blink leverages saliency-guided scanning and dynamic token resolution to selectively expand important visual tokens and drop less relevant ones, enabling adaptive visual perception. Comprehensive experiments on downstream tasks and further analyses validate the effectiveness of the framework and its consistent improvements, elucidating the underlying mechanisms and showing the soundness of our method.

\section*{Acknowledgement}

We would like to thank the anonymous reviewers for their comments, which greatly improve our paper. This work is supported by the National Natural Science Foundation of China (No. 62472419, 62472420) and the Enterprise Project (No. E4V06811F3).

{
    \small
    \bibliographystyle{ieeenat_fullname}
    \bibliography{main}
}

\clearpage
\setcounter{page}{1}
\maketitlesupplementary

\appendix

\section{Training Details}
\label{sec:appendix_train}

\subsection{Configurations}

For each involved transformer layer, we attach a lightweight token super-resolution (TokenSR) module to refine the expanded saliency tokens. Each TokenSR module consists of three sequential convolutional layers. The input and output channels are both set to 4096, matching the hidden dimension of the MLLM backbone, while the intermediate layers use 2048 and 1024 channels, respectively. The convolution kernel sizes are 5, 3, and 1, each applied with symmetric padding to preserve spatial resolution.\footnote{Code is available at: \url{https://github.com/yuchenblah/Blink}.}

\subsection{Datasets}

To train the TokenSR modules, we use the processed LLaVA-1.5~\cite{liu2024improved} training set, which is constructed from several widely used vision–language datasets. For each image, we additionally generate four quadrant crops (top-left, top-right, bottom-left, bottom-right) and record their positions relative to the original image. Each processed sample therefore consists of the full image, one cropped image, and the corresponding positional metadata of the cropped quadrant.

During training, both the full image and the cropped image are converted into pixel-value tensors and fed into the frozen MLLM backbone to obtain hidden representations. The cropped image provides the teacher features, while the TokenSR module takes as input the token hidden states corresponding to the same spatial region in the full image after bilinear interpolation, and generates enhanced token hidden states. The optimization objective is to align the enhanced token hidden states with the teacher features of the corresponding spatial region.

\subsection{Hyper-parameters}

\begin{table}[htbp]
    \centering
    % \small
    \begin{tabularx}{0.35\textwidth}{Xc}
        \toprule
        \textbf{Training Parameter} & \textbf{Value} \\
        \midrule
        \# GPUs & 2 \\
        Sequence length & 1024 \\
        Data type & Bfloat16 \\
        Learning rate & 1e-4 \\
        Learning rate scheduler & Cosine \\
        Warmup ratio & 0.03 \\
        Optimizer & AdamW \\
        Global batch size & 8 \\
        Epoch & 1 \\
        DeepSpeed & Zero-2 \\
        \bottomrule
    \end{tabularx}
    \caption{Training parameters used in the experiments.}
    \label{tab:parameters}
\end{table}

The training hyper-parameters for the TokenSR modules are presented in Tab.~\ref{tab:parameters}. 
All experiments are conducted on 8 H800 (80G) GPUs in total, where each TokenSR module is trained on 2 GPUs with a global batch size of 8, using BFloat16 precision. 
We set the warmup ratio to 3\% and employ a cosine learning rate scheduler, decaying the learning rate from 1e-4 to 0. 
Our implementation is based on HuggingFace Transformers~\cite{wolf2020transformers} and DeepSpeed~\cite{rasley2020deepspeed}.

\section{Implementation Details}
\label{sec:appendix_infer}

In our main experiments in Sec.~\ref{sec:exp}, the inference settings are configured as follows. For the layer ranges and saliency thresholds, we use layers 12–18 with $\tau_{\text{exp}} = 0.5$ and $\tau_{\text{drop}} = 0.4$ for MME and MM-Vet. For GQA, the same layer range is used with $\tau_{\text{exp}} = 0.6$ and $\tau_{\text{drop}} = 0.4$. For POPE, we apply layer 18 with $\tau_{\text{exp}} = 0.5$ and $\tau_{\text{drop}} = 0.4$. For MMBench, MMBench-CN and ScienceQA, we use the same layer with $\tau_{\text{exp}} = 0.25$.

Additionally, in the ablation studies on the MME benchmark, the High $\tau_{\text{exp}}$ / Low $\tau_{\text{drop}}$ setting corresponds to values of 0.7 and 0.3, respectively. The High $\tau_{\text{exp}}$ configuration refers to using $\tau_{\text{exp}} = 0.7$ while keeping $\tau_{\text{drop}}$ fixed at 0.4, consistent with the main experiment on MME.

\begin{table*}[t]
    \centering
    \small
    \setlength{\tabcolsep}{4pt}
    \renewcommand{\arraystretch}{1.2}
    \begin{tabular}{p{2.5cm} *{11}{c}}
        \toprule
        \multirow{2}{*}{\textbf{Method}} &
        \multicolumn{11}{c}{\textbf{MME$_{\text{Perception}}$}} \\
        \cmidrule(lr){2-12}
        & Exist. & Count & Pos. & Color & Poster & Celeb. & Scene & Landm. & Artw. & OCR & \textbf{Total} \\
        \midrule
        Vanilla & 195.00 & \textbf{158.33} & \textbf{135.00} & 175.00 & 150.68 & 146.47 & \textbf{164.25} & \textbf{167.50} & \textbf{134.75} & \textbf{102.50} & 1529.48 \\
        \rowcolor{blue!5}
        \textit{Blink-interp (Ours)} & 195.00 & \textbf{158.33} & \textbf{135.00} & \textbf{180.00} & 150.68 & \textbf{147.35} & \textbf{164.25} & \textbf{167.50} & 134.00 & \textbf{102.50} & \textbf{1534.62} \\
        \rowcolor{blue!5}
        \textit{Blink (Ours)} & \textbf{200.00} & \textbf{158.33} & 130.00 & \textbf{180.00} & \textbf{153.74} & 142.94 & 163.50 & 166.00 & 134.00 & \textbf{102.50} & 1531.02 \\
        
        \midrule
        
        \multirow{2}{*}{\textbf{Method}} &
        \multicolumn{5}{c}{\textbf{MME$_{\text{Cognition}}$}} & \multirow{2}{*}{\textbf{GQA}} & \multirow{2}{*}{\textbf{MMBench}} & \multirow{2}{*}{\textbf{MMBench$_{\text{CN}}$}} & \multirow{2}{*}{\textbf{POPE}} & \multirow{2}{*}{\textbf{SQA$_{\text{Img}}$}} & \multirow{2}{*}{\textbf{MM-Vet}} \\
        \cmidrule(lr){2-6}
        & CS & Num & Text & Code & \textbf{Total} & & & & & & \\
        \midrule
        Vanilla & \textbf{120.00} & \textbf{57.50} & 102.50 & \textbf{35.00} & 315.00 & 64.26 & \textbf{69.07} & 60.82 & 86.37 & 70.40 & 39.80 \\
        \rowcolor{blue!5}
        \textit{Blink-interp (Ours)} & \textbf{120.00} & 55.00 & 110.00 & \textbf{35.00} & 320.00 & 64.22 & \textbf{69.07} & 60.82 & 86.23 & \textbf{70.50} & 40.40 \\
        \rowcolor{blue!5}
        \textit{Blink (Ours)} & \textbf{120.00} & \textbf{57.50} & \textbf{117.50} & \textbf{35.00} & \textbf{330.00} & \textbf{64.29} & \textbf{69.07} & \textbf{60.91} & \textbf{86.40} & 70.45 & \textbf{40.50} \\
        \bottomrule
    \end{tabular}

    \caption{Downstream task performance across multiple benchmarks on LLaVA-NeXT. Vanilla denotes the base model, and our methods correspond to two configurations of Blink, where -interp indicates the variant that replaces the amplifier with training-free interpolation while retaining the Blink inference pipeline. The best scores are in \textbf{bold}.}
    \label{tab:downstream_next}
\end{table*}

\section{Evaluation on Other Backbones}
\label{sec:appendix_backbone}

\subsection{Evaluation on LLaVA-NeXT}

We evaluate Blink on the LLaVA-NeXT-7B~\cite{liu2024llavanext} backbone, a moderate variable-resolution MLLM. As shown in Tab.~\ref{tab:downstream_next}, our method consistently improves performance across a wide range of downstream benchmarks. Blink yields a notable gain of 15.00 on MME$_{\text{Cognition}}$ compared with the vanilla model and achieves the highest scores on GQA, MMBench-CN, POPE, and MM-Vet. Although its MME$_{\text{Perception}}$ score is slightly lower than that of Blink-interp, Blink shows stronger advantages on most tasks, suggesting that the fully trained TokenSR module strengthens fine-grained multimodal understanding.

Blink-interp achieves the best performance on MME$_{\text{Perception}}$ and improves over the backbone model on ScienceQA and MM-Vet, highlighting the complementary benefits of the dynamic inference pipeline. These results collectively demonstrate that Blink generalizes well to stronger backbones and provides robust gains across diverse downstream tasks.

% For POPE, MMBench, MMBench-CN, GQA, ScienceQA, and MM-Vet, we use layers 14--20 with $\tau_{\text{exp}} = 0.5$ and $\tau_{\text{drop}} = 0.4$, while for MME, we use layers 12--20 with thresholds $\tau_{\text{exp}} = 0.4$ and $\tau_{\text{drop}} = 0.3$. 

\subsection{Evaluation on Qwen2.5-VL}

\begin{table}[t]
    \centering
    \small
    \setlength{\tabcolsep}{3.8pt}
    \renewcommand{\arraystretch}{1.2}
    \resizebox{\linewidth}{!}{
    \begin{tabular}{l c c c c c}
        \toprule
        \textbf{Method} & \textbf{MME$_{\text{Perp.}}$} & \textbf{MME$_{\text{Cogn.}}$} & \textbf{GQA} & \textbf{POPE} & \textbf{SQA$_{\text{Img}}$} \\
        \midrule

        Vanilla & \underline{1638.16} & 598.57 & 58.14 & 87.63 & 80.81 \\
        \rowcolor{blue!5} \textit{Blink-interp (Ours)} & \underline{1638.16} & \underline{606.07} & \textbf{58.37} & \textbf{87.82} & \underline{81.06} \\
        \rowcolor{blue!5} \textit{Blink (Ours)} & \textbf{1645.65} & \textbf{608.57} & \underline{58.28} & \underline{87.75} & \textbf{81.21} \\

        \bottomrule
    \end{tabular}
    }
    \caption{Downstream task performance on Qwen2.5-VL-7B. Vanilla denotes the base model, and our methods correspond to two configurations of Blink, where -interp indicates the variant that replaces the amplifier with training-free interpolation while retaining the Blink inference pipeline. The best scores are in \textbf{bold}, and the second-best results are \underline{underlined}.}
    \label{tab:downstream_qwen}
\end{table}

To further validate the effectiveness of Blink on modern multimodal backbones, we conduct additional experiments on Qwen2.5-VL-7B~\cite{bai2025qwen2}, a recent and strong vision-language model. We evaluate our method across four benchmarks, including MME~\cite{fu2025mmecomprehensiveevaluationbenchmark}, GQA~\cite{hudson2019gqa}, POPE~\cite{li2023evaluating}, and ScienceQA~\cite{lu2022learn}. We consider both Blink and Blink-interp, where the trainable modules are replaced with bilinear interpolation while retaining the same inference pipeline.

As shown in Tab.~\ref{tab:downstream_qwen}, both Blink and Blink-interp consistently outperform the vanilla model across multiple benchmarks, demonstrating the effectiveness of our design. Specifically, Blink achieves the best performance on MME, improving MME$_{\text{Perp.}}$ from 1638.16 to 1645.65 and MME$_{\text{Cogn.}}$ from 598.57 to 608.57. It also attains the highest score on ScienceQA, indicating stronger multimodal reasoning ability. Meanwhile, Blink-interp achieves the best results on GQA of 58.37 and POPE of 87.82, showing that even a training-free variant can effectively enhance visual grounding and perception. Notably, Blink-interp also improves MME$_{\text{Cogn.}}$ to 606.07 and ScienceQA to 81.06, outperforming the vanilla baseline. These results demonstrate consistent improvements across backbones, highlighting the robustness and general applicability of our method.

\begin{table}[t]
    \centering
    % \footnotesize
    \small
    \setlength{\tabcolsep}{6pt}
    \renewcommand{\arraystretch}{1}
    \resizebox{1\linewidth}{!}{
    \begin{tabular}{p{1.6cm} ccccc}
        \toprule
        \textbf{Method} &
        \textbf{MME$_{\text{Total}}$} &
        \textbf{GQA} &
        \textbf{POPE} &
        \textbf{FLOPs} &
        \textbf{\#Fwd Passes} \\
        \midrule
        ViCrop & 1804.65 & 60.98 & \textbf{87.25} & 21.5T & 3 \\
        \rowcolor{blue!5} \textit{Blink (Ours)} & \textbf{1881.53} & \textbf{61.98} & 85.23 & \textbf{9.73T} & \textbf{1} \\
        \bottomrule
    \end{tabular}
    }
    \caption{\small{Downstream task performance and efficiency of different methods on LLaVA-1.5-7B. FLOPs denote the theoretical maximum computation per image. \#Fwd Passes indicates the number of full model inferences required for each input. The best scores are in \textbf{bold}.}}
    \label{tab:baselines}
\end{table}

\section{Comparison with ViCrop}
\label{sec:appendix_baseline}

Direct comparison with dynamic perception methods is non-trivial, as they typically require multiple forward passes and incur higher inference costs. To ensure a fair evaluation, we reproduce ViCrop~\cite{zhang2025mllms} on LLaVA-1.5-7B and report results on three benchmarks. As shown in Tab.~\ref{tab:baselines}, our method achieves better performance on MME and GQA, while maintaining comparable results on POPE. These results demonstrate that Blink consistently improves multimodal perception over prior methods.

In terms of efficiency, Blink significantly reduces computational overhead by lowering FLOPs from 21.5T to 9.73T (a 54.7\% reduction) and requiring only a single forward pass instead of three, thereby reducing inference latency. Meanwhile, the additional memory overhead compared to vanilla LLaVA-1.5-7B is minimal, with GPU usage increasing from 19.7 GB to 20.8 GB (a 1.1 GB increase) under FP16. Overall, Blink achieves a better balance between performance and efficiency, making it more suitable for practical deployment.

\section{Cases of Attention Redistribution}
\label{sec:appendix_case}

Following the analysis in Sec.~\ref{sec:exp_attn}, we conduct additional visualization experiments to examine how the tokens generated by the TokenSR module influence attention redistribution across transformer layers after expansion. As in the main experiments, expansion is applied at layer 12, and the newly generated tokens are preserved in all subsequent layers without being dropped.

\begin{figure*}[t]
  \centering
  \includegraphics[width=0.8\linewidth]{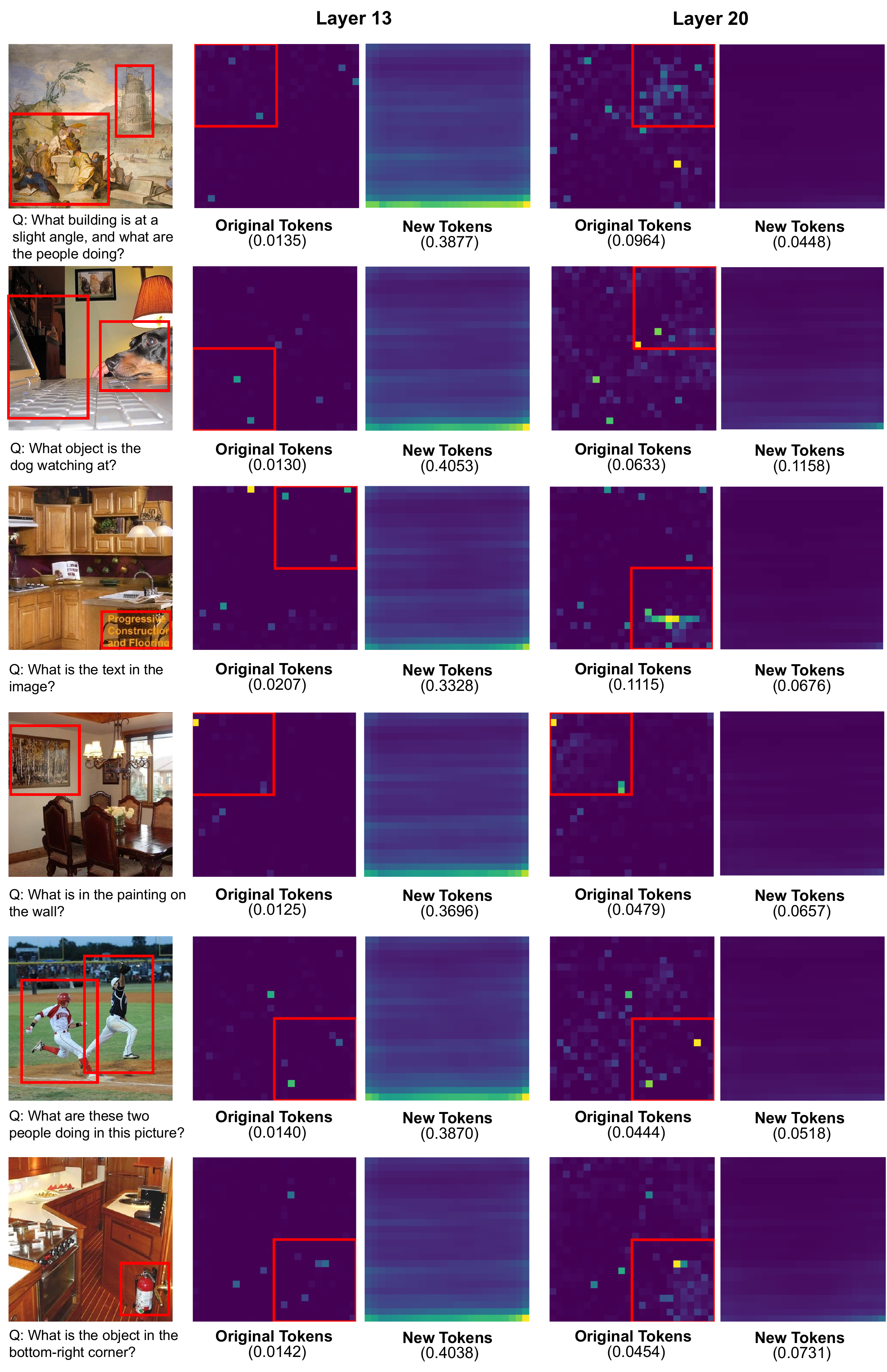}
  \caption{Visualization of attention redistribution after token expansion. Red boxes in the original image indicate the ground-truth important regions. The right panels show attention distributions on the original and expanded visual tokens at layers 13 and 20.}
  \label{fig:cases}
\end{figure*}

Fig.~\ref{fig:cases} presents attention maps across different layers to illustrate this spatial redistribution. We analyze the two examples from Sec.~\ref{sec:insights}, along with four additional cases. As expected, in all examples, attention on the newly introduced tokens remains more evenly distributed compared to the original visual tokens. Furthermore, in the first three cases, we observe clear layer-wise shifts in attention, consistent with our first key insight that the model progressively adjusts its attention allocation across layers after the expanded tokens are introduced.

\begin{table}[t]
    \centering
    \small
    \setlength{\tabcolsep}{6pt}
    \renewcommand{\arraystretch}{1.25}
    \begin{tabular}{l l c c c}
        \toprule
        \textbf{Method} & \textbf{\# Patches} & \textbf{Perc.} & \textbf{Cogn.} & \textbf{Total} \\

        \midrule

        \rowcolor{blue!5} \multirow{3}{*}{\textit{Blink-interp}} 
        & \textit{$2 \times 2$} & 1514.08 & 353.21 & 1867.29 \\
        & $3 \times 3$ & 1507.44 & 350.36 & 1857.80 \\
        & $4 \times 4$ & 1499.38 & 335.36 & 1834.74 \\

        \midrule

        \rowcolor{blue!5} \multirow{3}{*}{\textit{Blink}} 
        & \textit{$2 \times 2$} & 1519.74 & 361.79 & 1881.53 \\
        & $3 \times 3$ & 1498.16 & 374.64 & 1872.80 \\
        & $4 \times 4$ & 1500.24 & 358.21 & 1858.45 \\

        \bottomrule
    \end{tabular}
    \caption{Performance of Blink and its interpolation variant on the MME benchmark with different numbers of patches. The $2 \times 2$ setting is used in our main configuration, and the other configurations correspond to alternative patch partition settings.}
    \label{tab:patch_num}
\end{table}

\section{Ablation Study on Patch Numbers}
\label{sec:appendix_crop}

In our main experimental setup, the reshaped $H \times W$ attention grid used in saliency-guided scanning is uniformly partitioned into $2 \times 2$ patches. In this section, we evaluate the effect of varying the partition granularity by dividing the grid into $p \times p$ non-overlapping patches of equal size, and report results on the MME benchmark.

Since changing the number of patches alters the minimum proportion of the image that a salient region can occupy, we scale the saliency ratio thresholds accordingly. For the expansion threshold, the patch-adjusted value is computed as $\tau_{\text{exp}}^{(p)} = 0.5 \times \frac{1/p^{2}}{1/2^{2}}$, and similarly for the drop threshold, $\tau_{\text{drop}}^{(p)} = 0.4 \times \frac{1/p^{2}}{1/2^{2}}$.

As shown in Tab.~\ref{tab:patch_num}, neither the $3 \times 3$ nor $4 \times 4$ partitioning yields improvements over the default $2 \times 2$ setting. For Blink-interp, increasing the number of patches consistently degrades both perception and cognition scores, with the $4 \times 4$ configuration showing the largest decline. For Blink, the $3 \times 3$ variant offers a small gain in cognition but still reduces perception and overall MME performance, while the $4 \times 4$ setup leads to drops across all metrics. This may be because the current thresholds are derived using a simple proportional scaling rule, and more fine-grained tuning could further optimize performance for different patch granularities. Nevertheless, across all configurations, Blink equipped with the trained TokenSR module consistently outperforms Blink-interp, confirming that the learned amplifier is more effective than naive interpolation under all patch partitions.

\begin{table}[t]
    \centering
    \small
    \setlength{\tabcolsep}{6pt}
    \renewcommand{\arraystretch}{1.25}
    \begin{tabular}{l c c c}
        \toprule
        \textbf{Layers} & \textbf{Perc.} & \textbf{Cogn.} & \textbf{Total} \\
        \midrule

        \rowcolor{blue!5}
        \textit{Blink-interp (Ours)} & 1514.08 & 353.21 & 1867.29\\
        w/o interp & 1510.58\textsubscript{\textcolor[HTML]{d63031}{-3.50}} & 357.86\textsubscript{\textcolor[HTML]{0984e3}{+4.65}} & 1868.44\textsubscript{\textcolor[HTML]{0984e3}{+1.15}} \\
        
        \midrule

        \rowcolor{blue!5}
        \textit{Blink (Ours)} & 1519.74 & 361.79 & 1881.53\\
        w/o interp & 1515.08\textsubscript{\textcolor[HTML]{d63031}{-4.66}} & 355.00\textsubscript{\textcolor[HTML]{d63031}{-6.79}} & 1870.08\textsubscript{\textcolor[HTML]{d63031}{-11.45}} \\
        \bottomrule
    \end{tabular}
    \caption{Performance of Blink and its training-free varianton the MME benchmark with and without the interpolation step. In the w/o interp setting, the selected saliency tokens are inserted directly without being upsampled to match the original image resolution.}
    \label{tab:interp}
\end{table}

\section{Ablation Study on Interpolation}
\label{sec:appendix_crop}

In our method, prior to feeding saliency tokens into the TokenSR module, we first perform interpolation to upsample these tokens. The purpose of this interpolation is to align the length of the salient token sequence with that of the original image tokens, allowing both training and inference to proceed directly with standard two-image inputs at the original spatial resolution. This design ensures that the model operates on inputs that preserve the original spatial structure, which is more consistent with conventional visual encoding practices and facilitates effective learning of fine-grained multimodal representations.

Tab.~\ref{tab:interp} presents the performance on the MME benchmark with and without the interpolation step. In our experiments, the w/o interp condition refers to bypassing this initial upsampling. Specifically, after selecting the saliency tokens, they are directly extracted and inserted between the original visual and text tokens without resizing. For Blink-interp, w/o interp corresponds to inserting the saliency tokens directly into the dynamic inference pipeline without any training. For Blink, w/o interp means feeding the tokens into the trained TokenSR module without prior interpolation. From the experimental results, skipping interpolation for Blink-interp slightly reduces the perception score by 3.50, indicating that aligning the length of the expanded saliency sequence with the original image helps preserve spatial information. For Blink, removing interpolation leads to drops of 4.66 in perception, 6.79 in cognition, and 11.45 overall, further demonstrating that the TokenSR module benefits from receiving saliency tokens of the same length as the original image tokens, which allows the amplifier to more effectively refine multimodal representations. Overall, these results highlight that interpolation is a crucial step for fully leveraging spatial context and ensuring stable and accurate performance.

% \section{Count}
% \label{sec:appendix_count}

% \section{FLOPs}
% \label{sec:appendix_flops}

\section{Limitations and Future Work}
\label{sec:appendix_limit}

Although Blink demonstrates strong generalization across different tasks and backbones, its current design relies primarily on convolution-based upsampling. Exploring alternative expansion modules may further improve flexibility and performance.
Moreover, due to computational constraints, all experiments are conducted on models no larger than 7B parameters, leaving the scalability of Blink to larger MLLMs unverified.

Future work will explore alternative TokenSR mechanisms, such as multi-layer perceptrons, as complements or replacements for the convolution-based module, and investigate improvements that enhance the overall efficiency and practicality of Blink.

\end{document}